\documentclass[final]{cvpr}

\usepackage{times}
\usepackage{epsfig}
\usepackage{graphicx}
\usepackage{amsmath}
\usepackage{amssymb}
\usepackage{booktabs}
\usepackage{multirow}
\usepackage{subfigure} 
\usepackage{enumerate}
\usepackage{float}

\usepackage[pagebackref=true,breaklinks=true,colorlinks,bookmarks=false]{hyperref}

\def\cvprPaperID{****} 
\def\confYear{CVPR 2021}

\begin{document}

\title{LCCNet: LiDAR and Camera Self-Calibration using Cost Volume Network}

\author{Xudong Lv
\and
Boya Wang
\and
Ziwen Dou
\and
Dong Ye
\and
Shuo Wang
\and \\
School of Instrumentation Science and Engineering, Harbin Institute of Technology \\
{\tt\small 15B901019@hit.edu.cn, 15B901018@hit.edu.cn}
}
\maketitle
\thispagestyle{empty}
\pagestyle{empty}

\begin{abstract}
Multi-sensor fusion is for enhancing environment perception and 3D reconstruction in self-driving and robot navigation. Calibration between sensors is the precondition of effective multi-sensor fusion. Laborious manual works and complex environment settings exist in old-fashioned calibration techniques for Light Detection and Ranging (LiDAR) and camera. We propose an online LiDAR-Camera Self-calibration Network (LCCNet), different from the previous CNN-based methods. LCCNet can be trained end-to-end and predict the extrinsic parameters in real-time. In the LCCNet, we exploit the cost volume layer to express the correlation between the RGB image features and the depth image projected from point clouds. Besides using the smooth L1-Loss of the predicted extrinsic calibration parameters as a supervised signal, an additional self-supervised signal, point cloud distance loss, is applied during training. Instead of directly regressing the extrinsic parameters, we predict the decalibrated deviation from initial calibration to the ground truth. The calibration error decreases further with iterative refinement and the temporal filtering approach in the inference stage. The execution time of the calibration process is 24ms for each iteration on a single GPU. LCCNet achieves a mean absolute calibration error of $0.297cm$ in translation and $0.017{}^\circ $ in rotation with miscalibration magnitudes of up to $\pm 1.5m$ and $\pm 20{}^\circ $ on the KITTI-odometry dataset, which is better than the state-of-the-art CNN-based calibration methods. The code will be publicly available at https://github.com/LvXudong-HIT/LCCNet
\end{abstract}


\begin{figure}[thpb]
    \centering
    \includegraphics[width=7cm]{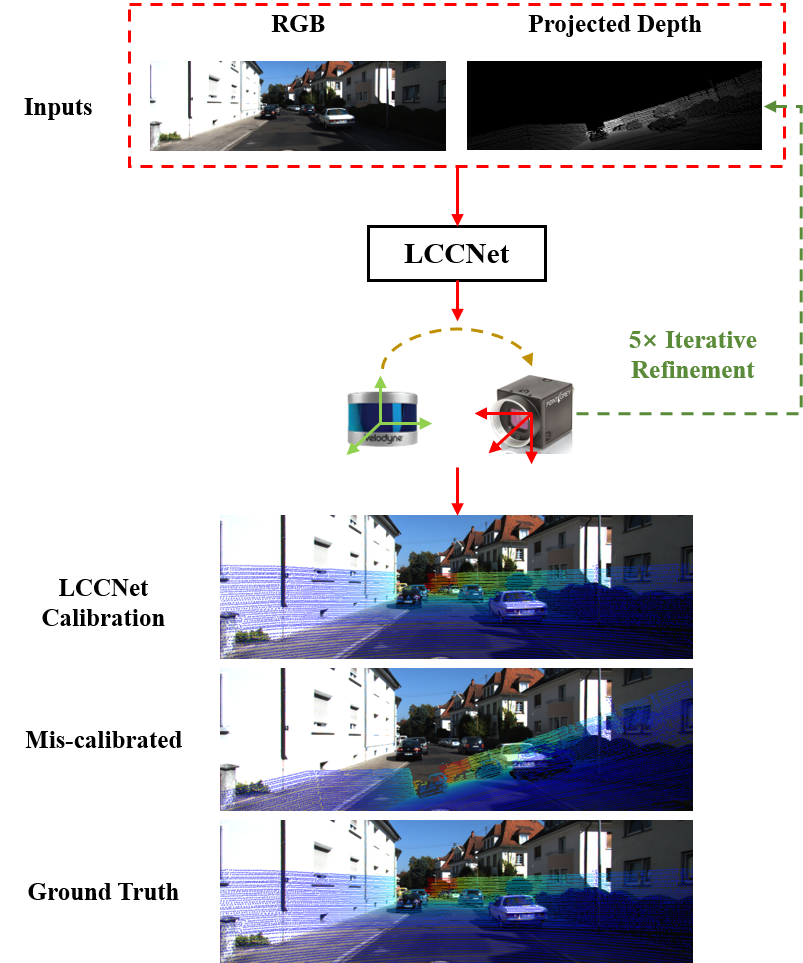}
    \caption{The proposed LCCNet takes the RGB and the projected depth image as inputs to predict the extrinsic parameters between the LiDAR and the camera. The point clouds are re-projected by the predicted extrinsic parameters. The re-projected depth image and the RGB image will be the subsequent inputs of the LCCNet. This process is called iterative refinement. After five times iterative refinements, we obtain the final extrinsic parameters estimation.}
    \label{LCCNet_show}
\end{figure}

\section{Introduction}

In the past few years, research on autonomous driving technology has developed rapidly. The working environment of autonomous driving is very complex and highly dynamic. No single sensor can ensure stable perception in all scenarios. The fusion of the LiDAR and the camera can provide accurate and stable perception for the surrounding environments or benefit for 3D reconstruction. The basis of LiDAR-camera fusion is the accurate extrinsic calibration, that is, precise estimation of the relative rigid body transformation. LiDAR point cloud and camera image belong to heterogeneous data. Point clouds are sparse 3D data, while images are dense 2D data. Calibration needs to accurately extract the 2D-3D matching correspondences between the pairs of a temporally synchronized camera and LiDAR frames.

Some early calibration works utilized artificial markers, such as checkerboard and specific calibration plates, to calibrate LiDAR and cameras. Most marker-based calibration algorithms are time-consuming, laborious, and offline, not suitable for self-driving car production. During the vehicle's operation, the position between the sensors will drift slightly with the running time. After a period of operation, the sensors need to be re-calibrated again to eliminate the accumulated error caused by drift. Some current calibration methods for LiDAR and cameras focus on fully automatic and target-less online self-calibration. However, most online self-calibration methods have strict requirements on calibration scenarios, and the calibration accuracy is not as high as offline calibration algorithms based on markers. Some researchers try to apply deep learning to the calibration tasks, using neural networks to predict the 6-DoF rigid body transformation between the two sensors. These methods directly fuse the features extracted from the image and the point cloud without considering the correlation between these two features.
e
This article proposes a new method for predicting extrinsic calibration parameters of an RGB camera and a LiDAR. More specifically, our contribution can be concluded as follows:
\begin{enumerate}[(1)]
    \item LCCNet is a novel end-to-end learning network for LiDAR-Camera extrinsic self-calibration. The network consists of three parts: feature extraction network, feature matching layer, and global feature aggregation network. We use the quaternion as the ground truth during training. Besides the smooth L1-Loss between the predicted calibration and ground truth, an additional point cloud distance loss is presented.
    \item The feature matching layer constructs a cost volume that stores the matching costs for RGB features and the corresponding Depth features. To our best knowledge, it is the first deep learning-based LiDAR-camera self-calibration approach that considering the correlation between features of different sensors.
    \item To further improve the calibration accuracy, iterative refinement and multiple frames analysis is applied. LCCNet has the best accuracy-speed trade-off compared to other state-of-the-art learning-based calibration methods.
\end{enumerate}


\section{Related Work}
The calibration between LiDAR and camera can be formulated as a 2D-3D registration problem to obtain the transformation of two sensors' coordinates. The calibration has three main solutions: (a) Target-based; (b) Target-less; (c) Learning-based.

\subsection{Target-based methods}
Geige et al. \cite{geiger2012automatic} realized LiDAR-camera calibration by multiple printed checkboard patterns on the walls and floors. Define a camera as a reference, all checkerboards in the target image are assigned to the nearest position in the reference image given the expectation. Wang et al. \cite{wang2017reflectance} developed a new automatic extrinsic calibration method for 3D LiDAR and panoramic camera using checkerboard. Multiple chessboards or auxiliary calibration object are utilized to provide extra 3D-3D or 2D-3D point correspondences \cite{kim2020extrinsic, an2020geometric}. 

To obtain more accurate 3D-3D or 2D-3D point correspondences, custom-made markers with a specific appearance, such as polygon plates, hollow circles, unmarked planes, and spherical targets, are introduced for calibration. Park et al. \cite{park2014calibration} estimated the corresponding 3D points by LiDAR scans on the edges of adjacent polygonal planar boards. The 2D vertical lines detected from the RGB image and the 3D vertical lines estimated from the LiDAR point cloud were regarded as calibration correspondences. \cite{guindel2017automatic} employed hollow circles as the calibration target to find the center of the circle in 2D image data and 3D LiDAR point cloud, respectively. Pusztai et al. \cite{pusztai2017accurate} adopted a carton of known size as the calibration plate to detect the box's surface. Mishra et al. \cite{9304750} proposed a LiDAR-camera extrinsic estimation algorithm on unmarked plane target by utilizing Planar Surface Point to Plane and Planar Edge Point to back-projected Plane geometric constraint. The surfaces and contours of the sphere can be accurately detected on point clouds and camera images, respectively. Thus, the spherical targets can achieve fast, and robust extrinsic calibration \cite{kummerle2018automatic, toth2020automatic}.

\subsection{Target-less methods}

Tamas et al. \cite{tamas2013targetless} proposed a nonlinear explicit correspondence-less calibration method regarding the calibration problem as a 2D-3D registration of a common LiDAR-camera region. Minimal information like depth data and shape of areas are applied to construct the nonlinear registration system, which directly provides the calibration parameters of the LiDAR-camera. Furthermore, \cite{tamas2014relative} advanced a new method of estimating 3D LiDAR and Omnidirectional Cameras' extrinsic calibration. Without using 2D-3D corresponding points or complex similarity measurement, this method depends on a set of corresponding regions and regresses the pose by solving a small nonlinear system of equations. Pandey et al. \cite{pandey2015automatic} adopted the effective correlation coefficient between the surface reflectivity measured by LiDAR and the intensity measured by a camera as one of the extrinsic parameters calibration function, while the other parameters remain unchanged.

Registering the gradient direction of the data obtained by LiDAR and camera was another target-less method. \cite{taylor2015multi} estimated the extrinsic parameter by minimizing the gradient's misalignment to realize data registration. Kang et al. \cite{kang2020automatic} employed the projection model-based edge alignment to construct the cost function, taking full advantage of the dense photometric and sparse geometry measurements.

\subsection{Learning-based methods}\label{sec:cnn}

RegNet \cite{schneider2017regnet} leverages the Convolutional Neural Networks (CNNs) to predict the 6-DoF extrinsic parameters between LiDAR and camera. CalibNet \cite{iyer2018calibnet} proposed a geometrically supervised deep network capable of automatically estimating the 6-DoF extrinsic parameters in real-time. The end-to-end training is performed by maximizing the geometric and photometric consistency between the input image and the point cloud. RGGNet \cite{yuan2020rggnet} utilized the Riemannian
geometry and deep generative model to build a tolerance-aware loss function.

Semantic information is introduced for obtaining an ideal initial extrinsic parameter. SOIC \cite{wang2020soic} transforms the initialization problem into the PNP problem of the semantic centroid. In this work, a matching constraint cost function based on the image's semantic elements and the LiDAR point cloud is presented. By minimizing the cost function, the optimal calibration parameter is obtained. Zhu et al. \cite{zhu2020online} regard extrinsic calibration as an optimization problem using semantic features to build a novel calibration quality metric.


\section{Method}\label{sec:Method}
\begin{figure*}[thpb]
    \centering
    \includegraphics[width=16cm]{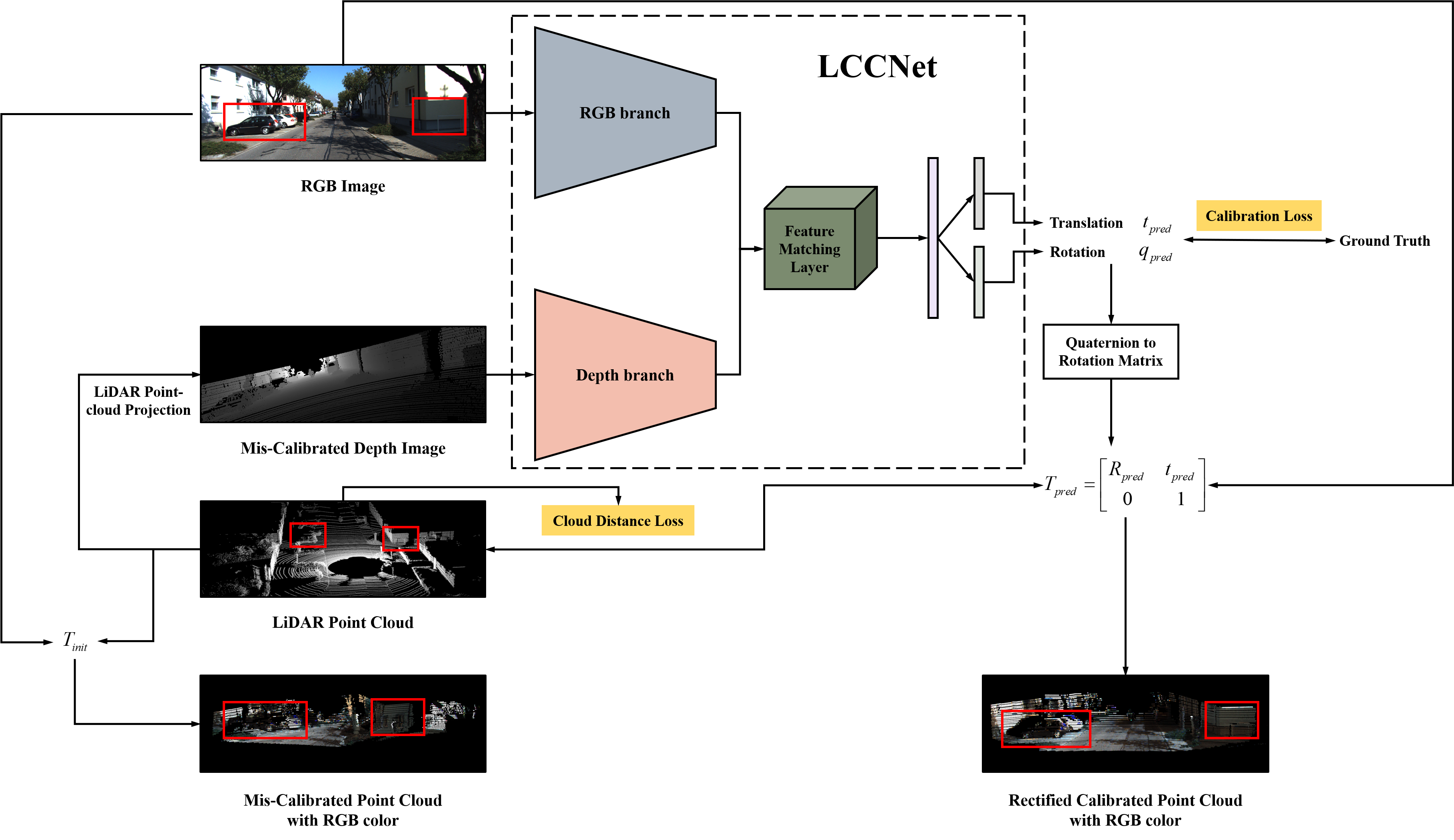}
    \caption{The workflow of our proposed method for the estimation of the extrinsic calibration parameters between 3D LiDAR and 2D camera. The network takes an RGB image from a calibrated camera and a projected sparse depth image from a mis-calibrated LiDAR as input. The output of the network is a 6-DoF rigid-body transformation ${{T}_{pred}}$ that represents the deviation between the initial extrinsic ${{T}_{init}}$ and the ground truth extrinsic ${{T}_{LC}}$. As shown, we notice that the 3D structure highlighted using red rectangles fails to project to their 2D counterparts with the mis-calibrated depth image. When using the predicted transformation ${{T}_{pred}}$ to revise the ${{T}_{init}}$, we can reconstruct a more consistent and accurate 3D scene structure.}
    \label{workflow block diagram}
\end{figure*}

We leverage CNNs to predict the 6-DoF extrinsic calibration between LiDAR and the camera. Our proposed method's workflow block diagram is shown in Figure \ref{workflow block diagram}.

\subsection{Input Processing}\label{sec:train data}
Given initial extrinsic ${{T}_{init}}$ and camera intrinsic $K$, we can generate the depth image by projecting each 3D LiDAR point cloud ${{P}_{i}}=[\begin{matrix}{{X}_{i}} & {{Y}_{i}} & {{Z}_{i}}\end{matrix}]\in {{\mathbb{R}}^{3}}$ from the LiDAR scan onto a virtual image plane with a 2D pixel coordinate ${{p}_{i}}=\left[ \begin{matrix}{{u}_{i}} & {{v}_{i}}\end{matrix} \right]\in {{\mathbb{R}}^{2}}$. The projection process is expressed as follows:
\begin{equation}
    \begin{aligned}
        Z_{i}^{init}\cdot \widehat{{{p}_{i}}}& ={Z_{i}^{init}}{{\left[ \begin{matrix}
           {{u}_{i}} & {{v}_{i}} & 1  \\
        \end{matrix} \right]}^{T}} \\
        & =K\left[ {{R}_{init}}|{{t}_{init}} \right]\widehat{{{P}_{i}}} \\ 
        & =K\left[ {{R}_{init}}|{{t}_{init}} \right]{{\left[ \begin{matrix}
           {{X}_{i}} & {{Y}_{i}} & {{Z}_{i}} & 1  \\
        \end{matrix} \right]}^{T}} \\ 
    \end{aligned}
\end{equation}

\begin{equation}
    {{T}_{init}}=\left[ \begin{matrix}
       {{R}_{init}} & {{t}_{init}}  \\
       0 & 1  \\
    \end{matrix} \right]
\end{equation}

where $\widehat{{{P}_{i}}}$ and $\widehat{{{p}_{i}}}$ represent the homogeneous coordinates of ${{P}_{i}}$ and ${{p}_{i}}$, ${R}_{init}$ and ${t}_{init}$ are the rotation matrix and translation vector of ${T}_{init}$. By using a Z-buffer method, the depth image ${{D}_{init}}$ is computed to determine the visibility of points along the same projection line, where every pixel $({{u}_{i}},{{v}_{i}})$ preserves the depth value $Z_{i}^{init}$ of a 3D point ${{P}_{i}}$ on camera coordinate.

\subsection{Network Architecture}\label{sec:network}
The proposed calibration network comprises three parts: feature extraction network, feature matching layer, and feature global aggregation. Since the parameters in each part are differentiable, CNN can be trained end-to-end. We will describe the structure and function of each part of this section.

\subsubsection{Feature Extraction Network}\label{sec:extraction}
The feature extraction network consists of 2 symmetric branches, extracting the RGB image and depth image features. For the RGB branch, we use a pre-trained ResNet-18 network, excluding the full connection layer. The architecture of the depth branch is consistent with the RGB branch. In the depth branch, we switch RELU to Leaky RELU as the activation function.

\subsubsection{Feature Matching Layer}\label{sec:matching}
After extracting features from two input modalities, a feature matching layer is adopted to calculate the matching cost for associating a pixel in RGB feature maps ${{x}_{rgb}}$ with its corresponding depth feature maps ${{x}_{lidar}}$. Inspired by PWC-Net \cite{sun2018pwc}, we take advantage of the correlation layer for feature matching. We define the constructed cost volume as the correlation between ${{x}_{rgb}}$ and ${{x}_{lidar}}$ that stores the matching cost:
\begin{equation}
    cv({{p}_{1}},{{p}_{2}})=\frac{1}{N}{{(c({{x}_{rgb}}({{p}_{1}})))}^{T}}c({{x}_{lidar}}({{p}_{2}}))
\end{equation}
where $c(x)$ is the flattened vector of feature maps $x$, $N$ is the length of the column vector $c(x)$, $T$is the transpose operator. For the features, we need to compute a local cost volume with a limited range of $d$ pixels, i.e.,${{\left| {{p}_{1}}-{{p}_{2}} \right|}_{\infty }}\le d$. Since the input feature maps are very small, $1/32$ of the full resolution images, we need to set the value $d$ very small ($d=2$ in this paper). The dimension of the 3D cost volume $cv$ is ${{d}^{2}}\times H\times W$, where $H$ and $W$ donate the height and width of feature maps ${{x}_{rgb}}$ and ${{x}_{lidar}}$, respectively.

\subsubsection{Feature Global Aggregation}\label{sec:aggregation}
LCCNet regresses the 6-DoF rigid-body transformation between LiDAR and camera with cost volume features. This network consists of a full connection layer with 512 neurons and two branches with stacked full connection layers representing rotation and translation. The output of the network is a $1\times 3$ translation vector. ${{t}_{pred}}$ and a $1\times 4$ rotation quaternion ${{q}_{pred}}$.

\subsection{Loss Function}\label{sec:loss}
Given an input pair composed of an RGB image $I$ and a depth image ${{D}_{init}}$, we use two types of loss terms during training: regression loss ${{L}_{T}}$ and point cloud distance loss ${{L}_{P}}$.
\begin{equation}
    L={{\lambda }_{T}}{{L}_{T}}+{{\lambda }_{P}}{{L}_{P}}
\end{equation}
where ${{\lambda }_{T}}$ and ${{\lambda }_{P}}$ denotes respective loss weight.

\subsubsection{Regression Loss}\label{sec:reg loss}
For the translation vector ${{t}_{pred}}$, the smooth $L1$ loss is applied. The derivative of $L1$ loss is not unique at zero, which may affect the convergence of training. Compared to $L1$ loss, the smooth $L1$ loss is much smoother due to the square function's usage near zero. Regarding the rotation loss ${{L}_{q}}$, since quaternions are essentially directional information, Euclidean distance cannot accurately describe the difference between the two quaternions. Therefore, we use angular distance to represent the difference between quaternions, as defined below:
\begin{equation}
    \begin{aligned}
        {{L}_{R}}={{D}_{a}}({{q}_{gt}},{{q}_{pred}})
    \end{aligned}
\end{equation}
where ${{q}_{gt}}$ is the ground truth of quaternion, ${{q}_{pred}}$ is the prediction, ${{D}_{a}}$ is the angular distance of two quaternions \cite{kendall2015posenet}. The total regression loss ${{L}_{T}}$ is the combination of translation and rotation loss:
\begin{equation}
    {{L}_{T}}={{\lambda }_{t}}{{L}_{t}}+{{\lambda }_{q}}{{L}_{R}}
\end{equation}
where ${L}_{t}$ is the smooth $L1$ loss for translation, ${{\lambda }_{t}}$ and ${{\lambda }_{q}}$ denotes respective loss weight.
\subsubsection{Point Cloud Distance Loss}\label{sec:pc loss}
Besides the regression loss, a point cloud distance constrain is added to the loss function. After transforming the quaternion ${{q}_{pred}}$ to a rotation matrix ${{R}_{pred}}$, we can obtain the homogeneous matrix ${{T}_{pred}}$:
\begin{equation}
    {{T}_{pred}}=\left[ \begin{matrix}
       {{R}_{pred}} & {{t}_{pred}}  \\
       0 & 1  \\
    \end{matrix} \right]
\end{equation}
Given a group of LiDAR point cloud $P=\left\{ {{P}_{1}},{{P}_{2}},\ldots ,{{P}_{N}} \right\},{{P}_{i}}\in {{\mathbb{R}}^{3}}$, the point cloud distance loss ${{L}_{P}}$ is defined as:
\begin{equation}
    {{L}_{p}}=\frac{1}{N}\sum\limits_{i=1}^{N}{{{\left\| T_{LC}^{-1}\cdot T_{pred}^{-1}\cdot {{T}_{init}}\cdot {{P}_{i}}-{{P}_{i}} \right\|}_{2}}}
\end{equation}
where ${T}_{LC}$ is the LiDAR-Camera extrinsic matrix, $N$ is the number of point clouds, ${{\left\| \cdot  \right\|}_{2}}$ denotes the $L2$ Normalization.

\subsection{Calibration Inference and Refinement}\label{sec:refinement}
The extrinsic calibration parameter between the uncalibrated LiDAR and camera can be obtained by combining the predicted results ${{T}_{pred}}$ of the calibration network and the initial calibration parameter ${{T}_{init}}$. The extrinsic calibration parameter is expressed as:
\begin{equation}
    {{\widehat{T}}_{LC}}=T_{pred}^{-1}\cdot {{T}_{init}}
\end{equation}

In this paper, we train multiple networks on different mis-calibration ranges. We choose the same translation and rotation range as \cite{schneider2017regnet}: $\left[ -x,x \right],x=\left\{ 1.5m,1.0m,0.5m,0.2m,0.1m \right\}$, $\left[ -y,y \right]$, $y=\left\{ 20{}^\circ ,10{}^\circ ,5{}^\circ ,2{}^\circ ,1{}^\circ  \right\}$. The latter smaller range is determined by the maximum mean absolute error (MAE) of the predicted results after training the network using the former larger range. The RGB image and the projected depth map will input the largest range $(\pm 1.5m,\pm 20{}^\circ)$ network. We regard the prediction ${{T}_{pred}}$  as ${{T}_{0}}$, and re-project the LiDAR point cloud with ${T_{0}^{-1}\cdot {{T}_{init}}}$ to generate a new depth image including more projected LiDAR points. The new depth image and the same RGB image will input to the second range $(\pm 1.0m,\pm 10{}^\circ)$ network to predict new transformation ${{T}_{1}}$. The aforementioned process is iterated five times to get the final extrinsic calibration matrix.
\begin{equation}
    {{\widehat{T}}_{LC}}={{\left( {{T}_{0}}\cdot {{T}_{1}}\ldots {{T}_{5}} \right)}^{-1}}\cdot {{T}_{init}}
\end{equation}

The calibration accuracy can be improved by using the multi-range network iterative refinement. We use the median of the predicted results of multiple frames as the final estimation of the extrinsic parameters. 


\section{Experiments and Discussion}\label{exp and dis}
We evaluate our proposed calibration approach on the KITTI odometry dataset. In this section, we detail the data preprocessing, evaluation metrics, training procedure and discuss the results of different experiments.
\subsection{Dataset Preparation}\label{data pre}
We use the odometry branch of the KITTI dataset \cite{geiger2012we} to verify our proposed algorithm. KITTI Odometry dataset consists of 21 sequences from different scenarios. The dataset provides calibration parameters between each sensor, among which the calibration parameters between LiDAR and camera were obtained by \cite{geiger2012automatic} as the ground truth of extrinsic calibration parameters. In this paper, we only consider the calibration between the LiDAR and the left color camera. Specifically, we used sequences from 01 to 20 for train and validation (39011 frames) and sequence 00 for test (4541 frames). The test dataset is spatially independent of the training dataset, except for a very small subset sequence (about 200 frames), so it can be assumed that the test scenario is not in the training data.

To solve the insufficiency of training data, we add a random deviation $\Delta T$ within a reasonable range to the extrinsic calibration matrix of LiDAR and camera. In this paper, we define the extrinsic parameter ${{T}_{LC}}$ as the Euclidean transformation from the LiDAR coordinate to the camera coordinate. After adding the random parameter $\Delta T$ to ${{T}_{LC}}$, the initial extrinsic ${{T}_{init}}=\Delta T\cdot {{T}_{LC}}$ is obtained. By randomly changing the deviation value, we can acquire a large amount of training data.

\subsection{Evaluation Metrics}\label{metrics}
The experimental results are analyzed according to the rotation and translation of the calibration parameters. The Euclidian distance between the vectors evaluates the translation vector. The absolute error of the translation vector is expressed as follows:
\begin{equation}
    {{E}_{t}}={{\left\| {{t}_{pred}}-{{t}_{gt}} \right\|}_{2}}
\end{equation}
where ${{\left\| \cdot  \right\|}_{2}}$ denotes the 2-norm of a vector. We also test the translation vector's absolute error in $X,Y,Z$ directions, respectively.

Quaternions represent the rotation part. Since quaternion means direction, we use quaternion angle distance to illustrate the difference between quaternions. To test the extrinsic rotation matrix's angle error on three degrees, we need to transform the rotation matrix to Euler angles and compute the angle error of Roll, Pitch, and Yaw.

\subsection{Training Details}\label{training details}
During the training stage, we use Adam Optimizer with an initial learning rate $3{{e}^{-4}}$. We train our proposed calibration network on two Nvidia GP100 GPU with batch size 120 and total epochs 120. For the multi-range network, it is not necessary to retrain each network from scratch. Instead, a large-range model can be regarded as the pre-trained model for small-range training to speed up the training process. The model's training epoch with the largest range is set to 120, while the others are set to 50.

\subsection{Results and discussion}\label{results}

\begin{figure*}[thpb]
    \centering
        \subfigure[]{\includegraphics[width=5.5cm]{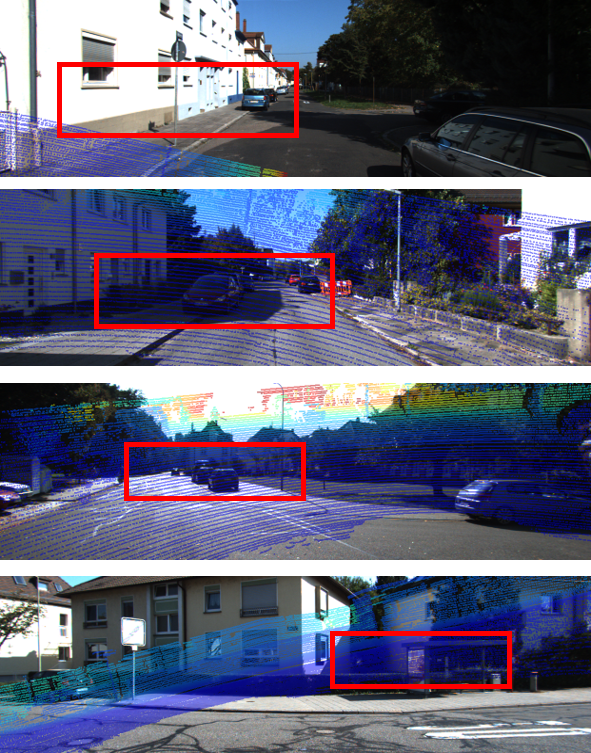}} 
        \subfigure[]{\includegraphics[width=5.5cm]{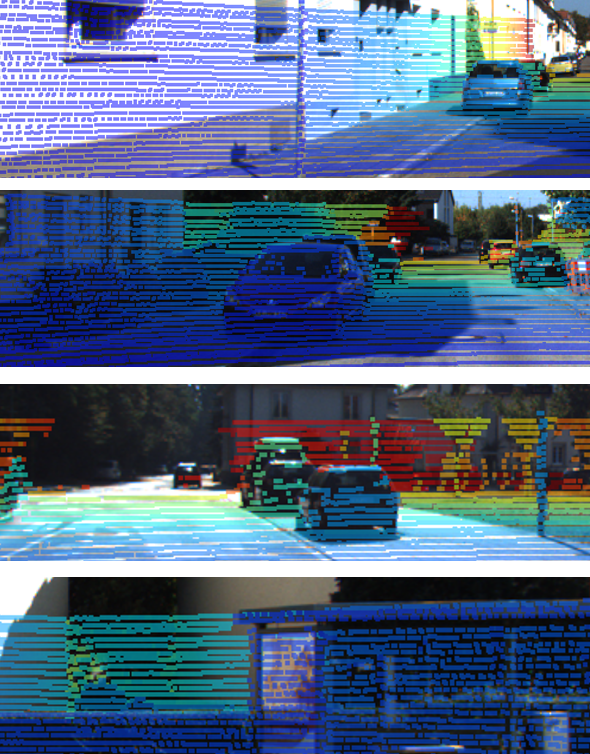}}
        \subfigure[]{\includegraphics[width=5.5cm]{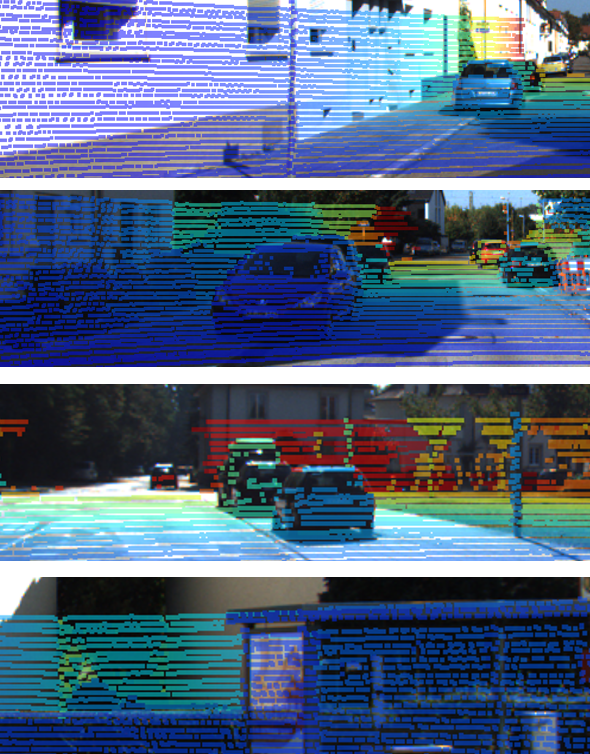}}
    \caption{Results of different single-shot calibration after multi-range iteration on the test dataset. (a) Initial Calibration, (b) Ground truth, (c) Calibration results.}
    \label{f9}
\end{figure*}

The visual results of the multi-range iterations are shown in Figure \ref{f9}. The final calibration results are shown in Table \ref{multi-range table}. It is obvious that after multi-range iterations, the calibration error is further reduced and the error distribution is concentrated at a smaller value. Our approach achieves a mean square translation error of 1.588cm, a mean translation error of 0.361cm (x, y, z: 0.243cm, 0.380cm, 0.459cm), and a mean quaternion angle error of $0.163{}^\circ $, a mean angle error of $0.030{}^\circ$(roll, pitch, yaw: $0.030{}^\circ ,\text{ }0.019{}^\circ ,\text{ }0.040{}^\circ $).

\begin{table*}[]
\caption{The results of the multi-range network iteration}
\label{multi-range table}
\begin{center}
\begin{tabular}{cccccccccc}
\toprule
\multicolumn{2}{c}{\multirow{2}{*}{Results of multi-range}}                 & \multicolumn{4}{c}{Translation (cm)} & \multicolumn{4}{c}{Rotation (${}^\circ   $)} \\
\multicolumn{2}{c}{}                                                        & ${{E}_{t}}$ & X      & Y     & Z     & ${{E}_{R}}$    & Roll    & Pitch   & Yaw     \\
\midrule
\multirow{3}{*}{After $20{}^\circ   /1.5\operatorname{m}$ network} & Mean   & 17.834         & 11.849         & 5.169          & 7.613          & 1.235          & 0.170          & 0.676          & 0.594          \\
                                                                   & Median & 13.423         & 14.415         & 5.337          & 4.188          & 0.898          & 0.033          & 0.615          & 0.549          \\
                                                                   & Std.   & 17.165         & 5.783          & 3.847          & 6.492          & 1.427          & 0.260          & 0.580          & 0.216          \\
\multirow{3}{*}{After $10{}^\circ   /1.0\operatorname{m}$ network} & Mean   & 6.291          & 2.045          & 4.195          & 2.238          & 0.594          & 0.378          & 0.391          & 0.405          \\
                                                                   & Median & 5.662          & 1.929          & 3.066          & 2.437          & 0.452          & 0.271          & 0.156          & 0.218          \\
                                                                   & Std.   & 3.494          & 1.629          & 3.673          & 1.683          & 0.637          & 0.422          & 0.480          & 0.453          \\
\multirow{3}{*}{After $5{}^\circ   /0.5\operatorname{m}$ network}  & Mean   & 3.915          & 1.267          & 2.212          & 1.107          & 0.414          & 0.309          & 0.330          & 0.334          \\
                                                                   & Median & 3.712          & 1.390          & 2.410          & 1.057          & 0.297          & 0.028          & 0.026          & 0.071          \\
                                                                   & Std.   & \textbf{0.809} & 0.686          & 0.989          & 0.570          & 0.541          & 0.507          & 0.538          & 0.479          \\
\multirow{3}{*}{After $2{}^\circ   /0.2\operatorname{m}$ network}  & Mean   & 2.069          & 0.664          & 0.633          & \textbf{0.281}          & 0.288          & 0.132          & 0.103          & 0.081          \\
                                                                   & Median & 1.592          & 0.475          & 0.455          & \textbf{0.316}          & 0.215          & 0.038          & 0.040          & 0.045          \\
                                                                   & Std.   & 1.859          & 0.497          & 0.581          & \textbf{0.112}          & 0.465          & 0.196          & 0.140          & 0.099          \\
\multirow{3}{*}{After $1{}^\circ   /0.1\operatorname{m}$ network}  & Mean   & \textbf{1.588} & \textbf{0.243} & \textbf{0.380} & 0.459 & \textbf{0.163} & \textbf{0.030} & \textbf{0.019} & \textbf{0.040} \\
                                                                   & Median & \textbf{1.011} & \textbf{0.262} & \textbf{0.358} & 0.352 & \textbf{0.121} & \textbf{0.030} & \textbf{0.001} & \textbf{0.022} \\
                                                                   & Std.   & 1.776          & \textbf{0.053} & \textbf{0.253} & 0.254 & \textbf{0.435} & \textbf{0.019} & \textbf{0.031} & \textbf{0.039}  \\
\bottomrule                                                     
\end{tabular}
\end{center}
\end{table*}

\begin{figure*}[thpb]
    \centering
        \subfigure[Initial calibration]{\includegraphics[width=17cm]{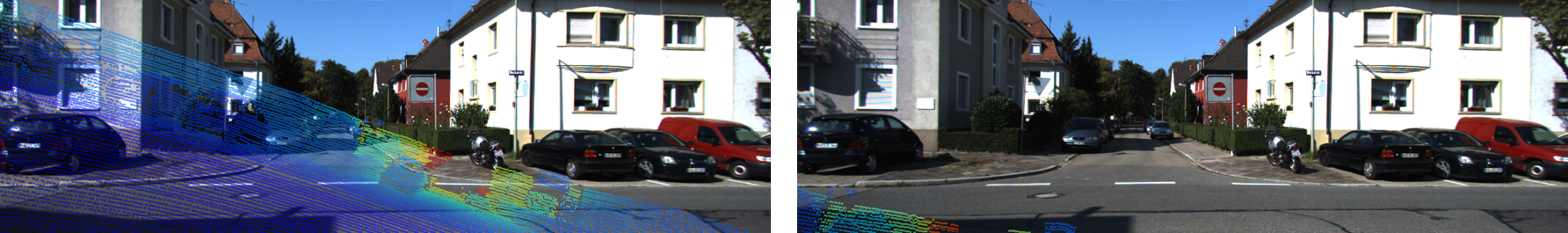}} \\ 
        \subfigure[Calibration results]{\includegraphics[width=17cm]{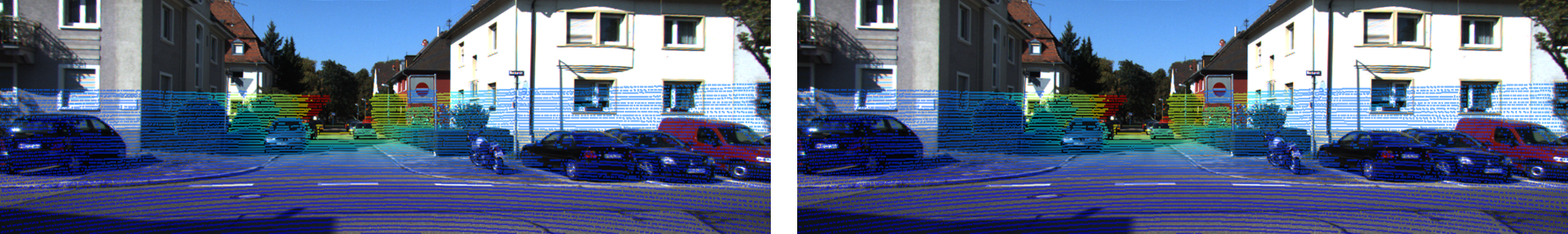}} \\
        \subfigure[Calibration error with temporal filtering]{\includegraphics[width=17cm]{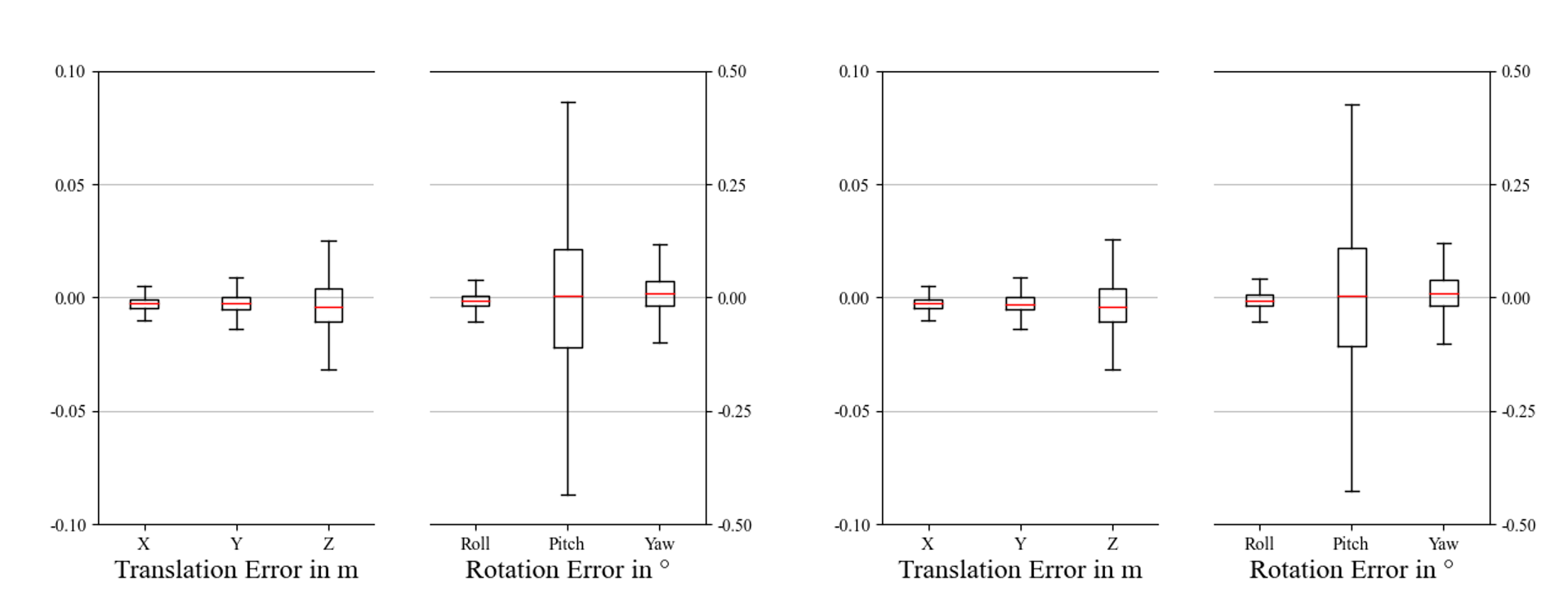}} \\
    \caption{Examples of the error distribution for a miscalibration, which is fixed over the test sequence. Five networks are executed iteratively ($20{}^\circ /1.5m,10{}^\circ /1.0m,5{}^\circ /0.5m,2{}^\circ /0.2m,1{}^\circ /0.1m$).}
    \label{f10}
\end{figure*}

From Figure \ref{f10}, we can find that our proposed calibration method can accurately predict the extrinsic calibration parameters for different random initial parameters. Although there is a great difference between the two initial parameters shown in Figure \ref{f10}(a), their corresponding calibration results are almost in full accord. Figure \ref{f10}(c) also shows that the two tests' error distributions are very similar. The proposed method has a high tolerance for the deviation of initial extrinsic parameters; that is, the algorithm can perform the calibration task accurately even a few point clouds are projected into the image. The results of the multiple frames analysis are exhibited in Table \ref{tf table}, the algorithm achieves a mean square translation error of 1.010cm, a mean translation error of 0.297cm (x, y, z: 0.262cm, 0.271cm, 0.357cm) and a mean quaternion angle error of $0.122{}^\circ $, a mean angle error of $0.017{}^\circ$(roll, pitch, yaw: $0.020{}^\circ ,\text{ }0.012{}^\circ ,\text{ }0.019{}^\circ $). The comparison results with other learning-based extrinsic calibration methods in Table \ref{comparison table} express that our proposed method is superior to these state-of-the-art algorithms. Due to most of the training datasets are the same, we do not re-train the baselines.

\begin{table*}[ht]
\caption{The results of the calibration error with multiple frames analysis}
\label{tf table}
\begin{center}
\begin{tabular}{ccccccccc}
\toprule
\multirow{2}{*}{Results} & \multicolumn{4}{c}{Translation (cm)} & \multicolumn{4}{c}{Rotation (${}^\circ $)}  \\
                         & ${{E}_{t}}$     & X      & Y     & Z     & ${{E}_{R}}$    & Roll  & Pitch & Yaw   \\
\midrule
Mean                                                            & 1.010        & 0.262 & 0.271 & 0.357 & 0.122          & 0.020   & 0.012   & 0.019   \\
Median                                                          & 1.010        & 0.260 & 0.274 & 0.337 & 0.122          & 0.020   & 0.010   & 0.019   \\
Std.                                                            & 0.007        & 0.087 & 0.109 & 0.148 & 0.001          & 0.008   & 0.007   & 0.009   \\
\bottomrule
\end{tabular}
\end{center}
\end{table*}

\begin{table*}[ht]
\caption{Comparison results with other learning-based calibration algorithms.}
\label{comparison table}
\begin{center}
\begin{tabular}{cccccccccc}
\toprule
\multirow{2}{*}{Method} & \multirow{2}{*}{Mis-calibrated range}       & \multicolumn{4}{c}{Translation absolute Error (cm)} & \multicolumn{4}{c}{Rotation absolute Error (${}^\circ $)} \\
                        &                                             & mean        & X           & Y          & Z          & mean         & Roll         & Pitch        & Yaw          \\
\midrule
Regnet \cite{schneider2017regnet}         & $[-1.5m,1.5m]/[-20{}^\circ ,20{}^\circ ]$ & 6           & 7           & 7          & 4          & 0.28         & 0.24         & 0.25         & 0.36         \\
Calibnet \cite{iyer2018calibnet}       & $[-0.2m,0.2m]/[-10{}^\circ ,10{}^\circ ]$ & 4.34        & 4.2         & 1.6        & 7.22       & 0.41         & 0.18         & 0.9          & 0.15         \\
Ours                    & $[-1.5m,1.5m]/[-20{}^\circ ,20{}^\circ ]$ & \textbf{0.297} & \textbf{0.262} & \textbf{0.271} & \textbf{0.357} & \textbf{0.017} & \textbf{0.020} & \textbf{0.012} & \textbf{0.019}        \\
\bottomrule
\end{tabular}
\end{center}
\end{table*}

\section{Conclusion}\label{conclusion}
We propose a novel learning-based extrinsic calibration method for the 3D LiDAR and the 2D camera. The calibration network consists of three parts: feature extraction network, feature matching layer and feature global aggregation network. We construct a cost volume between the RGB features and the depth features for feature matching instead of concatenating them directly compared to other learning-based approaches. To deal with the training data's insufficiency, we add a random deviation to the extrinsic transformation matrix. Therefore, the network does not predict the extrinsic parameters between LiDAR and camera directly, but the random deviation. Besides the extrinsic ground truth's supervision, we also add a point cloud constrain to the loss function. Iteratively refinement with multi-range networks and multiple frames analysis will further decrease the calibration error. Our method achieves a mean absolute calibration error of $0.297cm$ in translation and $0.017{}^\circ $ in rotation with miscalibration magnitudes of up to $\pm 1.5m$ and $\pm 20{}^\circ $, which is superior to other state-of-the-art learning-based methods.

{\small
\bibliographystyle{ieee_fullname}
\bibliography{egbib}
}

\end{document}